\documentclass[3p]{elsarticle}
\usepackage{hyperref}
\usepackage{booktabs}
\usepackage{amsmath}
\usepackage{algorithm}
\usepackage{algpseudocode}
\usepackage[english]{babel}
\usepackage[utf8]{inputenc}
\usepackage{color}
\usepackage{graphicx}
\usepackage{multirow}
\usepackage{makecell}
\usepackage{adjustbox}
\usepackage{todonotes}

\begin{document}

\begin{frontmatter}

\title{Explaining Machine Learning Predictive Models through Conditional Expectation Methods}

\author[itiaddress]{Silvia Ruiz-España\corref{equal}}
\ead{silviaruiz@iti.es}

\author[itiaddress]{Laura Arnal\corref{equal}}
\ead{larnal@iti.es}

\author[itiaddress]{François Signol}
\ead{fsignol@iti.es}

\author[itiaddress]{Juan-Carlos Perez-Cortes}
\ead{jcperez@iti.es}

\author[itiaddress]{Joaquim Arlandis}
\ead{arlandis@iti.es}

\address[itiaddress]{ITI, Universitat Polit\`ecnica de Val\`encia, Camino de Vera, s/n, 46022 Val\`encia, Spain}

\cortext[equal]{These authors contributed equally to this work.}

\begin{abstract}
The rapid adoption of complex artificial intelligence (AI) and machine learning (ML) models has led to their characterization as black boxes due to the difficulty of explaining their internal decision-making processes. This lack of transparency hinders users’ ability to understand, validate and trust model behavior, particularly in high-risk applications. Although explainable AI (XAI) has made significant progress, there remains a need for versatile and effective techniques to address increasingly complex models. 
This work introduces Multivariate Conditional Expectation (MUCE), a model-agnostic method for local explainability designed to capture prediction changes from feature interactions. MUCE extends Individual Conditional Expectation (ICE) by exploring a multivariate grid of values in the neighborhood of a given observation at inference time, providing graphical explanations that illustrate the local evolution of model predictions. In addition, two quantitative indices, stability and uncertainty, summarize local behavior and assess model reliability. Uncertainty is further decomposed into $uncertainty^+$ and $uncertainty^-$ to capture asymmetric effects that global measures may overlook. 
The proposed method is validated using XGBoost models trained on three datasets: two synthetic (2D and 3D) to evaluate behavior near decision boundaries, and one transformed real-world dataset to test adaptability to heterogeneous feature types. Results show that MUCE effectively captures complex local model behavior, while the stability and uncertainty indices provide meaningful insight into prediction confidence. MUCE, together with the ICE modification and the proposed indices, offers a practical contribution to local explainability, enabling both graphical and quantitative insights that enhance the interpretability of predictive models and support more trustworthy and transparent decision-making.	
\end{abstract}


\end{frontmatter}

\section{Introduction}
\label{sec:intro}

Artificial Intelligence (AI) has experienced significant advancements in recent years, enabling machines to perform complex tasks and make decisions that resemble human intelligence \cite{Ali2024}. These advancements have driven widespread adoption of AI and machine learning models across various sectors, including web search engines, speech recognition, autonomous vehicles, strategy games, image analysis, medical procedures, and national defense, among others \cite{Mersha2024,Samek2017,Weller2019}.

As AI techniques proliferate, their implementations have exceeded even the highest expectations across many domains \cite{Shrivastava2023}. As AI tackles increasingly difficult problems, the methodologies used have become increasingly complex \cite{Mersha2024}. Most of these models, particularly Deep Learning (DL) models, are inherently complex and lack explanations for the decision-making processes, leading them to be described as black boxes \cite{Hassija2024}. This term refers to machine learning models that operate as opaque systems, where the internal operations of the model are not easily accessible or interpretable \cite{Garg2021}. This lack of transparency makes it difficult for users to understand model behavior, identify potential biases or errors, or hold the model accountable for its decisions \cite{Hassija2024}. In response to these concerns, explainable AI (XAI) has emerged as a critical area of research for improving the transparency and interpretability of AI systems. XAI focuses on developing AI systems that not only produce accurate results but also provide explanations for their decisions in ways understandable to humans. This transparency is crucial for fostering trust in AI systems, especially in high-risk applications such as healthcare, finance, and autonomous systems. By understanding how AI systems reach their conclusions, users can verify the reliability of the results, identify potential biases, and take corrective actions if necessary \cite{Ali2024,Hassija2024}. Furthermore, XAI significantly improves AI system performance, particularly in design, debugging, and decision-making processes \cite{Samek2017,Samek2019,Zhang2018}.

Although interest in XAI has steadily increased, there is still a lack of consensus regarding symbolism and terminology. Therefore, contributions to this field often depend on the terminology and theoretical framework specific to each researcher \cite{Marcus2018,Mersha2024}. With this in mind, various explainability techniques have been developed to address the needs of different users, problems, and behaviors. The work of Hassija \textit{et al.} \cite{Hassija2024} provides a useful categorization of explanation types, emphasizing the goal of interpreting the logic behind black-box algorithms. In this categorization, techniques are classified into three main categories: stage-based (ante-hoc or post-hoc), scope-based (global or local), and example-based. This structure allows users to select the most appropriate approach according to their explainability needs.  

Ante-hoc and post-hoc explainability techniques are two distinct approaches to explaining the internal mechanisms of AI systems. The key difference between them lies in the stage of implementation \cite{Guidotti2018}. Ante-hoc XAI techniques are employed during training and development, typically by limiting the complexity of the machine learning model to make it more transparent and interpretable \cite{Molnar2022}. Such models ---sometimes called intrinsically explainable models, transparent, or glass-box models \cite{Retzlaff2024}--- have interpretability built into the model itself due to their simpler structure, allowing users to understand how decisions are made based on the data. Common examples include linear regression models \cite{Molnar2022}, logistic regression \cite{Molnar2022}, decision trees \cite{Molnar2022}, decision rules \cite{Molnar2022}, rule-based learners \cite{Hassija2024}, and Bayesian Rule Lists \cite{Hassija2024,Letham2015}. Due to their ease of providing explanations, interpretable ante-hoc methods have always been preferred over other black-box methods. However, accuracy may be affected as a consequence of this intrinsic interpretability \cite{Breiman2001}. 

The primary reason for using post-hoc procedures is the challenge of achieving an optimal balance between accuracy and interpretability \cite{Sarkar2016}. Post-hoc explanation techniques are applied after AI models have already been trained, to clarify model predictions or decision-making processes. These approaches are generally categorized into two main types: model-specific and model-agnostic methods \cite{Hassija2024}. While model-specific explainability techniques are valuable, they provide only a limited range of insights for interpreting predictions from black-box models. To address this limitation, model-agnostic methods were developed, providing interpretation tools that work independently of the model type. Their universal applicability is achieved by analyzing both input and output data simultaneously, although they cannot access internal model details, such as weights or key parameters. Recently, there has been an increased focus on developing these model-agnostic methods to broaden the scope of XAI, highlighting the increasing need for versatile explanatory tools across diverse AI models and application areas. 

When focusing on scope, explainability methods can be categorized into two main types: global explanations, which provide an overall understanding of how the model functions, and local explanations, which help interpret specific decisions made by the model. Global explainability approaches aim to clarify the general logic of a model, its behavior across the entire input space, and to identify the features that are most significant to the model overall. These approaches can be further classified into model-extraction and feature-based methods. The model extraction approach involves training an interpretable model, such as a linear model or a decision tree, to mimic the decisions made by a black-box model. This process produces what is often referred to as a global surrogate model. Within this approach, Rule Extraction and Model Distillation are two of the most popular strategies \cite{Hassija2024}. However, it is important to note that these methods may also compromise accuracy due to the oversimplification of model complexity. To maintain the desired level of accuracy, methods that analyze the impact or importance of input features on algorithms have been explored. In addition to examining feature interactions \cite{Molnar2022,Friedman2008}, feature-based methods also study the individual importance of each input feature in contributing to model predictions. This analysis enables a deeper understanding of which features are most relevant and how they impact the outcome, even in complex models. Partial Dependency Plots (PDP) and Accumulated Local Effects (ALE) plots are among the most widely used techniques for evaluating feature importance. It is worth noting that both methods are model-agnostic \cite{Hassija2024,Molnar2022}. These approaches offer a global perspective on how features influence model predictions without oversimplifying their complexity. In addition, network representations based on feature interactions and dependencies \cite{ChowLiu} provide a graphical insight into feature relationships, complementing feature‑based methods by highlighting how variables jointly contribute to model behavior.

Local explainability methods focus on understanding individual predictions and clarifying why a model made a specific decision for a given instance. Among the most widely recognized techniques are Individual Conditional Expectation (ICE) plots \cite{Goldstein2015}, which visually show how variations in a specific feature influence predictions for individual instances. This approach serves as a local counterpart to PDP, which provides a broader overview across the entire dataset. One of the most commonly used methods is Local Interpretable Model-Agnostic Explanations (LIME) \cite{Ribeiro2016a}. LIME approximates the predictions of any classifier by constructing a simpler, interpretable model that faithfully represents the complex model behavior around the specific prediction. It works by perturbing the input data to observe changes in the model predictions and then fitting an interpretable model, such as linear regression, to these altered data observations, providing a localized and straightforward interpretation of the underlying complex model. Another valuable technique is Anchors \cite{Ribeiro2018}, which establishes ‘anchor’ rules for individual predictions, outlining the necessary conditions under which the prediction remains consistent when certain features change. Additionally, SHapley Additive exPlanations (SHAP) \cite{Lundberg2017} applies game theory concepts to quantify the contribution of each feature to the prediction. SHAP calculates the Shapley value, reflecting the expected marginal contribution of each feature while accounting for all possible combinations of features. Developing local model-agnostic techniques is crucial for understanding individual predictions from black-box models, which enhances trust in automated decision-making processes, especially in critical areas like healthcare. These techniques also help identify and mitigate biases in AI systems, promoting fairness and ensuring the responsible and ethical use of AI across diverse applications \cite{Letrache2023}. 

Regarding example-based explainability methods, these approaches utilize specific instances from the dataset to clarify the behavior of machine learning models or to explain the underlying distribution of the data. Generally, these methods show notable effectiveness when instances have significant context in their feature values, which is often the case with images or structured text. Commonly used example-based methods include counterfactual explanations, adversarial examples, prototypes and criticisms, as well as influential instances. Counterfactual explanations explore how modifying an instance could alter the model's output, while adversarial examples reveal potential vulnerabilities in the model's decision boundaries. Prototypes identify representative cases, whereas criticisms highlight atypical instances. Influential instances, on the other hand, highlight data observations that have had the most significant impact on the model training process \cite{Molnar2022}.

It is worth mentioning that XAI has evolved as AI applications increasingly lean towards DL. As a result, to address the challenges associated with the complexity of neural networks, specific techniques and methodologies have been developed to attempt to reveal the decision-making processes within these sophisticated models. These techniques are essential to ensure transparency and interpretability in key DL applications, such as image recognition \cite{Pak2017}, natural language processing \cite{Young2018}, or complex pattern recognition tasks \cite{Bai2021}, among others. The work by Hosain \textit{et al.} \cite{Hosain2024} provides a useful categorization of XAI techniques for DL, organizing them based on their methodologies and applications in the field of DL interpretation. They classify these techniques into six main categories: propagation-based techniques, sequence and text understanding, gradient-based visualization techniques, generative models for explanations, model-agnostic and surrogate modeling, and concept-based interpretation. However, the inherent complexity of DL architectures still presents significant challenges in providing clear and interpretable explanations for model decisions \cite{Saeed2023}. Despite the progress made, a pressing need remains for further research and development of effective techniques to address these challenges.  

Therefore, local explanations are crucial as they offer insights into individual predictions, which is particularly valuable in high-risk decision-making scenarios. Furthermore, model-agnostic methods that can be applied to DL architectures are increasingly important as these models continue to dominate various AI applications due to their superior performance. In this context, our work introduces a novel XAI technique that addresses these specific needs, offering a local, model-agnostic approach that is also applicable to DL models that work with tabular data. This technique not only provides details about the model behavior in the local context of each observation but also offers indices that allow the evaluation of two crucial aspects: the stability of features in that specific environment and the degree of uncertainty in the decision made by the model, a combined perspective on behavior and local reliability.

\section{Materials and methods}
\label{sec:matnmet}

\subsection{Datasets}
\label{sec:dataset}

To support the development and validation of the proposed algorithms, synthetic datasets were generated to create controlled environments that enable precise evaluation and assessment. 
This approach facilitates the generation of diverse contexts, including homogeneous regions populated exclusively by observations of the same class, sparsely populated regions, and edge cases with observations situated near the decision boundaries between classes.
The use of simulated data across various scenarios provides an ideal setting to test the robustness and effectiveness of the proposed methodology. Additionally, a real-world dataset was employed for two primary purposes: first, to adapt the developed algorithms so that they can be applied to all types of features; second, to provide diverse examples of practical use in a more realistic context. 

The following subsections provide a detailed description of the three datasets used in this study: two synthetic and one real-world. In all cases, machine learning models were trained to perform binary classification tasks (see Section \ref{sec:results}), which is essential for demonstrating how the proposed methods can provide insights into model decision-making.

\subsubsection{Synthetic 2D dataset}
\label{sec:dataset_2d}

\begin{figure}[H]
	\centering
	\includegraphics[width=0.55\textwidth]{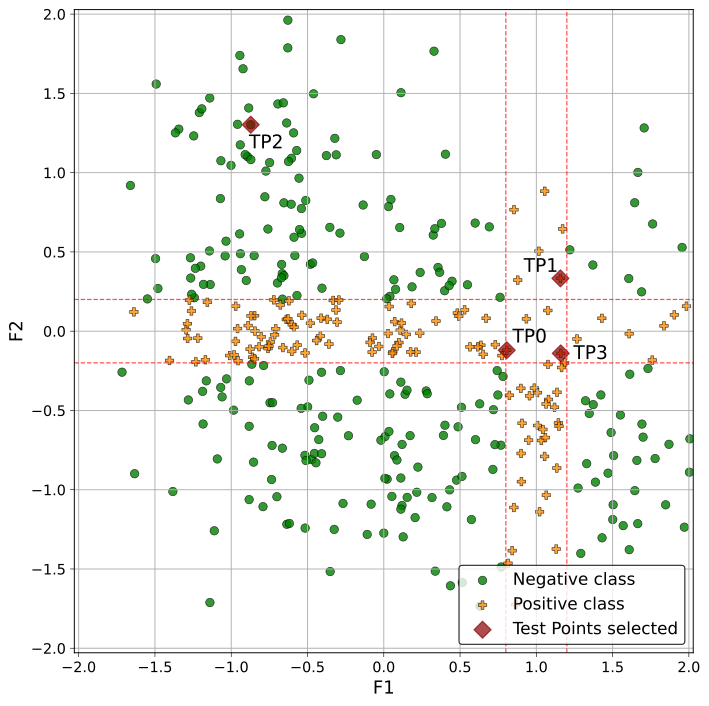}
	\caption{Cross-like scenario. 2D synthetic dataset with 400 observations and two features (\textit{F1}, \textit{F2}). The negative class is represented by green circles and the positive class by orange crosses. The red dashed line represents the class boundary. The dark red diamonds indicate four representative data observations selected to be explained with the XAI methods proposed in Sections~\ref{sec:ice} and \ref{sec:muce}.}
	\label{fig:dataset_synthetic2D}
\end{figure}

The first generated dataset is composed of two features (\textit{F1}, \textit{F2}). In this scenario, all positive class observations are located inside a cross-like structure. Observations corresponding to the negative class are distributed throughout the rest of the space, creating a boundary between the two classes that defines the overall shape, as shown in Figure \ref{fig:dataset_synthetic2D}. 

The dataset contains 400 observations, with 132 (33\%) belonging to the positive class. Figure \ref{fig:dataset_synthetic2D} also highlights four representative data observations (TP0, TP1, TP2, and TP3), which will be used to validate the proposed algorithms and indices presented in this work.

\subsubsection{Synthetic 3D dataset}
\label{sec:dataset_3d}

\begin{figure}[H]
	\centering
	\includegraphics[width=0.65\textwidth]{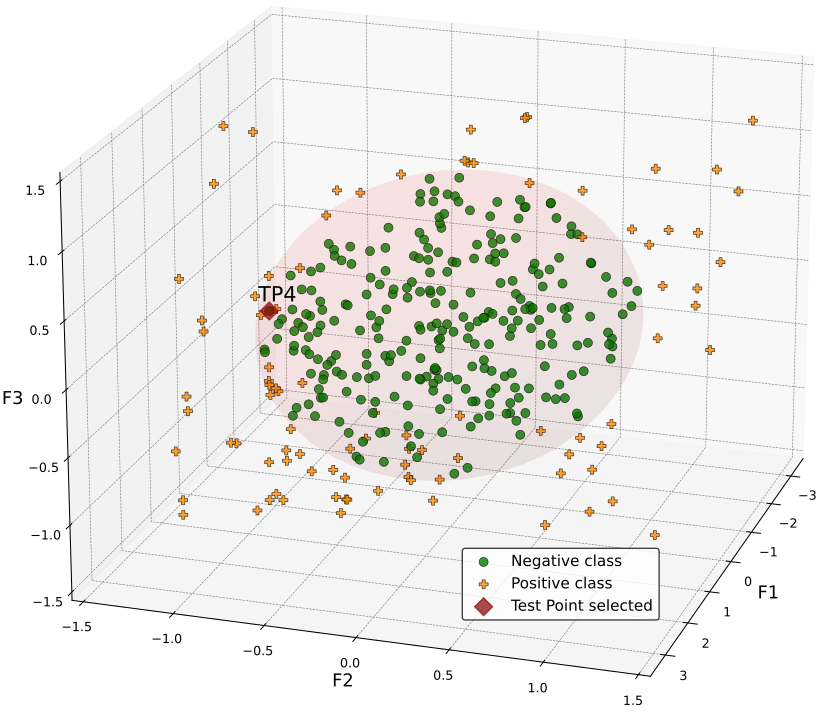}
	\caption{Ellipsoidal scenario: 3D synthetic dataset with 400 observations and three features (\textit{F1}, \textit{F2}, \textit{F3}). The negative class is represented by green circles and the positive class by orange crosses. The pink surface represents the ellipsoidal class boundary. The dark red diamond indicates the representative data observation to be explained with the XAI methods proposed in Sections~\ref{sec:ice} and \ref{sec:muce}.}
	\label{fig:dataset_synthetic3D}
\end{figure}

The second generated dataset is a three-feature scenario (\textit{F1}, \textit{F2}, \textit{F3}), allowing for the exploration of non-linear interactions. This dataset has been generated to extend the validation by gradually increasing dimensionality from 2D to 3D, while retaining visual readability. In this scenario, the negative class observations are enclosed within an ellipsoid defined by radii $x=3$, $y=1$, and $z=1$. Positive class observations are distributed outside this ellipsoidal boundary, resulting in a clear three-dimensional separation between the two classes, as shown in Figure \ref{fig:dataset_synthetic3D}. 

The dataset contains 400 observations, with 132 (33\%) belonging to the positive class. Figure \ref{fig:dataset_synthetic3D} also highlights a representative data observation (TP4) located close to the ellipsoidal boundary. This observation will serve as a reference case for illustrating the behavior and validating the proposed explanation techniques.

\begin{figure}[H]
	\centering
	\includegraphics[width=1\textwidth]{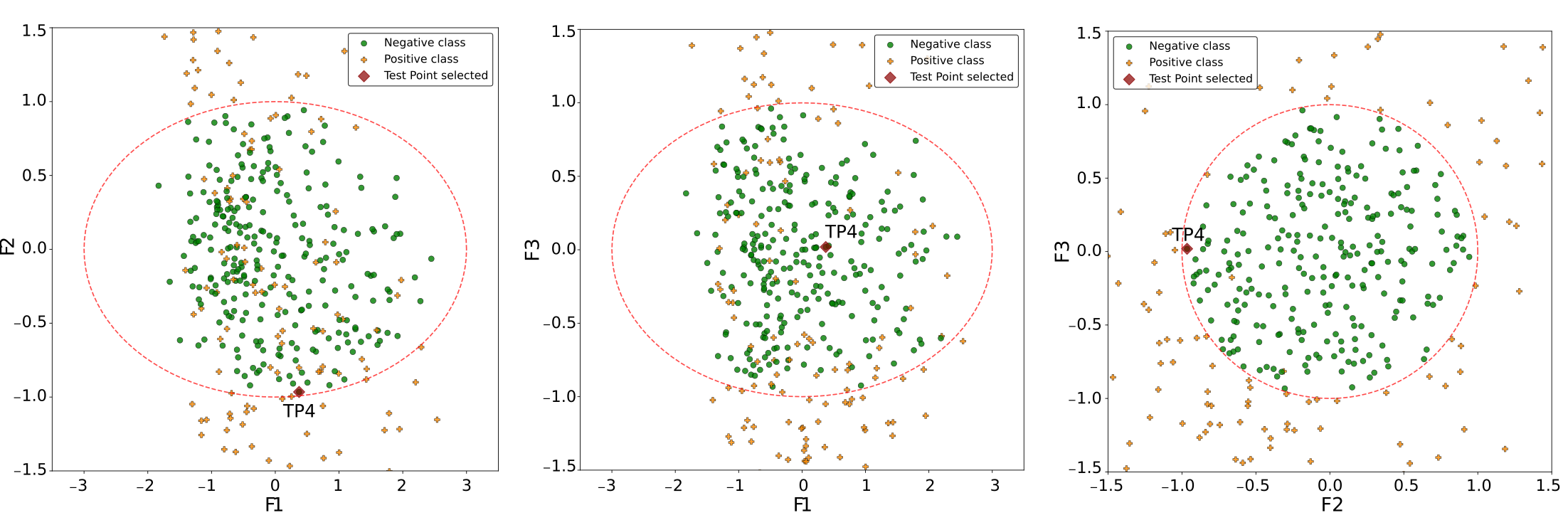}
	\caption{2D projections of the synthetic 3D dataset (ellipsoidal scenario): (a) XY plane projection (F1-F2), (b) XZ plane projection (F1-F3), and (c) YZ plane projection (F2-F3). The negative class is represented by green circles and the positive class by orange crosses. The red dashed line represents the projected class boundary. The dark red diamond indicates the representative data observation selected for further analysis.}
	\label{fig:dataset_synthetic3D_projections}
\end{figure}

Due to the three-dimensional nature of the dataset, the 3D visualization may lead to misinterpretation regarding the spatial relationship between data observations and the ellipsoidal boundary. Therefore, we include 2D projections onto the XY, XZ, and YZ planes (Figure \ref{fig:dataset_synthetic3D_projections}) to provide a complementary perspective. While these projections may show positive class observations appearing to fall inside the ellipsoid due to dimensionality reduction, they help clarify the spatial distribution and the position of the selected reference observation.

\subsubsection{Real-world dataset}
\label{sec:dataset_real}

To demonstrate the applicability of the proposed XAI methodology to problems of any dimensionality, the next validation dataset is a real, higher-dimensional one. The selected real-world dataset is based on information collected by the U.S. Census Bureau in 1990, where each row contains data corresponding to a census block group. A block group is the smallest geographical unit for which the U.S. Census Bureau publishes sample data, typically including a population of 600 to 3000 people \cite{CaliforniaHousing_Pace1997}. 

Originally, this dataset was intended for a regression task, where the target feature represents the median house value in California districts, expressed in hundreds of thousands of U.S. dollars (\$100,000). However, for this study, the problem was reformulated as a binary classification task by using the median value of the target feature as a threshold. Observations with a median house value below the overall median were labeled as 0 (low-value housing), while those above the median were labeled as 1 (high-value housing). 

The dataset consists of 20,640 observations and eight features. It is verified that the class distribution is nearly balanced, with 50.01\% of instances labeled as low-value housing and 49.99\% as high-value housing. The features used in this dataset are described in Table \ref{tab:real_dataset_features}.

\begin{table}[h]
	\centering
	\caption{Description of the features in the original real-world dataset.}
	\begin{tabular}{lllc}
		\hline
		\textbf{\textit{Feature}} & \textbf{\textit{Description}} & \textbf{\textit{Type}}  & \textbf{\textit{Median [5th - 95th percentile]}} \\
		\hline
		\textit{medinc}     & Median income in block group                      & Float & 3.53 [1.60 - 7.30] \\
		\textit{houseage}   & Median house age in block group                   & Integer & 29 [8 - 52]\\
		\textit{averooms}   & Average number of rooms per household             & Float & 5.23 [3.43 - 7.64]\\
		\textit{avebedrms}  & Average number of bedrooms per household          & Float & 1.05 [0.94 - 1.27]\\
		\textit{population} & Block group population                            & Integer & 1166 [348 - 3288]  \\
		\textit{aveoccup}   & Average number of household members               & Float & 2.82 [1.87 - 4.33]\\
		\textit{latitude}   & Block group latitude                              & Float & 34.26 [32.82 - 38.96]\\
		\textit{longitude}  & Block group longitude                             & Float & -118.49 [-122.47 - -117.08]\\
		\hline
	\end{tabular}
	\label{tab:real_dataset_features}
\end{table}

In order to demonstrate the model's validity in a general case, a preprocessing step was performed to remove outlier observations.
The purpose of this filter is to mitigate the influence of extreme values, which can unnecessarily complicate the model's learning of class boundaries. The analysis of the proposed method's behaviour in sparsely populated regions of space is a task for future work.
Specifically, for each feature taken independently, values below the 5th percentile or above the 95th percentile were considered outliers. All observations containing at least one outlier value in any of the features were removed. As a result of this filtering process, the dataset was reduced to 9,490 observations out of the original 20,640.

Furthermore, as previously mentioned, another objective of using this real-world dataset is to handle different types of features and to provide practical examples in a more realistic setting. To this end, some of the original features from the preprocessed dataset were modified to introduce greater heterogeneity in feature types, including binary, ordinal, and categorical formats. 

To create a binary feature, the original integer feature \textit{population} was used, following the same approach applied to the target feature when converting it into a classification problem. Specifically, a new feature \textit{population\_bin} was created by computing the median value of the feature, and all observations with a population below the median were assigned a value of 0 (low population). In contrast, those above the median were assigned a value of 1 (high population).

To create an ordinal feature, the original feature \textit{medinc}, which contains information about the median income per block group, was used. Based on the minimum and maximum values of this feature, as well as its overall distribution, the data were divided into five categories. Each category was then assigned a numerical value, ranging from 0 (lowest income group) to 4 (highest income group), thereby converting the feature into an ordinal feature (\textit{medinc\_ord}).

\begin{figure}[h]
	\centering
	\includegraphics[width=0.7\textwidth]{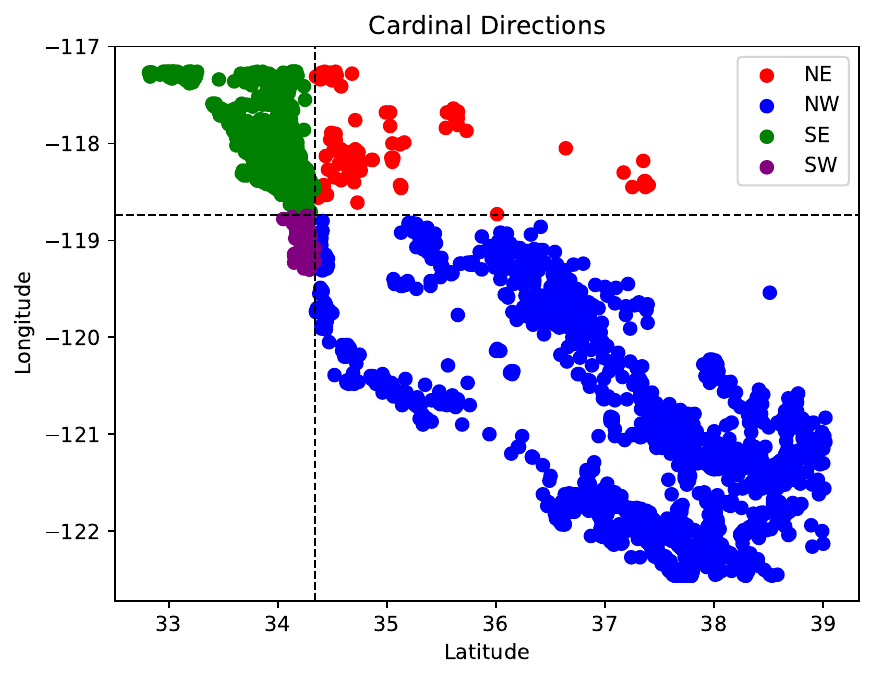}
	\caption{Spatial distribution of the observations according to their assigned cardinal direction, derived from latitude and longitude. The dashed lines represent the median values used to divide the geographic space into four quadrants.}
	\label{fig:cardinal_directions}
\end{figure}

To generate a categorical feature, a new feature representing cardinal directions was derived from the original \textit{latitude} and \textit{longitude} features. The median values of both coordinates were computed to divide the geographic space into four quadrants: Northeast (NE), Northwest (NW), Southeast (SE), and Southwest (SW). Each observation was assigned to one of these categories depending on whether its latitude and longitude values were above or below their respective medians, hence creating the new categorical feature (\textit{cardinal\_point}). Figure \ref{fig:cardinal_directions} displays the spatial distribution of the observations according to their assigned cardinal direction, with each category represented by a distinct color.

After these transformations, the dataset includes a more heterogeneous set of features, incorporating integer, continuous, binary, ordinal, and categorical features. The original features were removed to avoid including multicollinearity and reduce redundancy. Table \ref{tab:final_dataset_features} summarizes the final set of seven features.

\begin{table}[h]
	\centering
	\caption{Final set of features in the real-world dataset after feature transformations.}
	\label{tab:final_dataset_features}
	\begin{tabular}{lll}
		\hline
		\textbf{\textit{Feature}} & \textbf{\textit{Description}} & \textbf{\textit{Type}} \\
		\hline
		\textit{houseage}     & Median house age in block group                     & Integer \\
		\textit{averooms}     & Average number of rooms per household              & Float \\
		\textit{avebedrms}    & Average number of bedrooms per household           & Float \\
		\textit{aveoccup}     & Average number of household members                & Float \\
		\textit{population\_bin} & Binary indicator of high (1) or low (0) population & Binary \\
		\textit{medinc\_ord}  & Ordinal income category (0: lowest to 4: highest)  & Ordinal \\
		\textit{cardinal\_point} & Geographic quadrant (NE, NW, SE, SW)              & Categorical \\
		\hline
	\end{tabular}
\end{table}

\subsection{Modified Individual Conditional Expectation (ICE)}
\label{sec:ice}

The Individual Conditional Expectation (ICE) plot~\cite{Goldstein2015} is a visualization tool for interpreting supervised learning models, building upon the concept of Partial Dependence Plots (PDP)~\cite{friedman2001greedy}. In essence, this method illustrates how the model's predictions change when varying the values of a subset of features while keeping all other features constant.
In short, for a given observation and a given feature $F_{i} \in F$, the value is substituted for each of the other values that are present in the dataset for that feature $F_{i}$, keeping all other features unmodified. For each replacement of a value, the model prediction is computed. One curve is then generated for each observation by plotting the values of $F_{i}$ against the corresponding model predictions.
The original paper~\cite{Goldstein2015} generalizes the method to subsets of features. However, in practice, only one or two features are usually changed simultaneously due to difficulties in plotting and posterior analysis.

Originally, the authors proposed ICE as an exploratory tool rather than a method for local explainability. Consequently, both the original implementation~\cite{Goldstein2015} and widely used libraries such as scikit-learn~\cite{scikit-learn} require multiple observations to construct ICE plots. To the best of our knowledge, these implementations do not support generating a single curve in isolation, as plots are typically produced in batches.

To address this limitation, we implemented a custom ICE method that utilizes a configurable grid of values for each feature, spanning from the minimum to the maximum observed values in the training dataset. These minimum and maximum values are stored during training, allowing ICE curves to be generated at inference time without access to the full dataset. This methodology enables the generation of ICE plots for individual observations. The pseudocode for this procedure is presented in Algorithm~\ref{code:ice}.

The resulting plot also highlights the feature value corresponding to the specific observation, facilitating the analysis of prediction behavior near that value.

\begin{algorithm}[H]
\caption{Pseudocode for the modified ICE}
\label{code:ice}
\begin{algorithmic} 
\Function{FitICE}{$X, n$}
	\Comment{$X$: training data, $n$: no. of observations to generate for the grid}
	\State $G \gets zeros(size(columns(X)))$
	\For{$col \in columns(X)$}
		\State $v_{min} \gets min(X[col])$
		\State $v_{max} \gets max(X[col])$
		\State $G[col] \gets linspace(v_{min}, v_{max}, n)$
		\Comment{n evenly spaced observations between $v_{min}, v_{max}$}
	\EndFor
	\State \textbf{return} $G$
\EndFunction\\

\Function{ComputeICE}{$G, \mathbf{x}, F_{i}, model$}
	\Comment{$G$: learned data grid, $\mathbf{x}$: observation to test, $F_{i}$: feature to test, $model$: estimator to explain}
	\State $O_{ice} \gets \emptyset$
	\State $\hat{y}_{ice} \gets \emptyset $
	\For{$v \in G[F_{i}]$}
		\State $\mathbf{x^{'}} \gets \mathbf{x}$
		\State $\mathbf{x^{'}}[F_{i}] \gets v$
		\State $O_{ice} \gets O_{ice} \cup v$
		\State $\hat{y}_{ice} \gets \hat{y}_{ice} \cup model(\mathbf{x^{'}})$
	\EndFor
	\State \textbf{return} $O_{ice}, \hat{y}_{ice}$
\EndFunction
\end{algorithmic}
\end{algorithm}

\subsection{Multivariate Conditional Expectation (MUCE)}
\label{sec:muce}

\begin{figure}[!hbtp]
	\centering
	\includegraphics[width=0.8\textwidth]{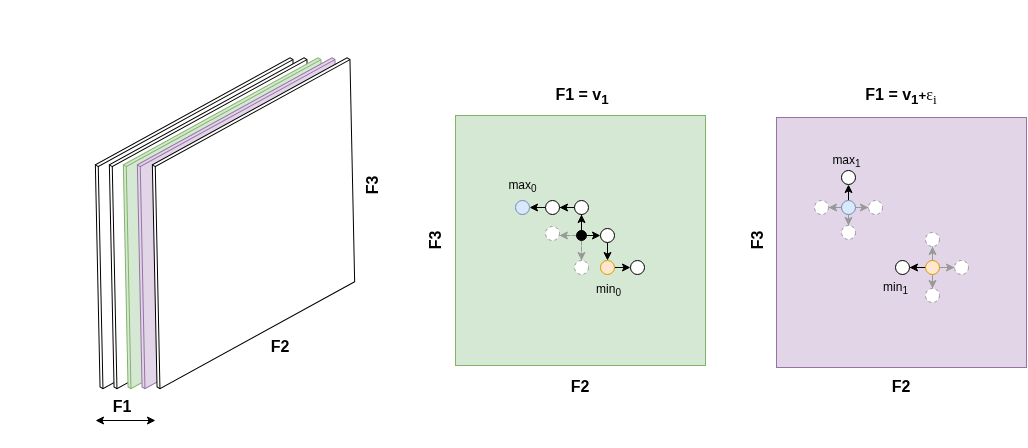}
	\caption{MUCE algorithm overview for three features: $F_{i}=F1$ and $F_{j}=\{F2, F3\}$. The exploration begins at the observation (black circle) in the first iteration (green plane), and observations are generated over $F_{j}$ (white circles). Among the newly generated values, the maximum and minimum predicted values are selected (solid circles), and the other generated values are discarded (dashed circles). This exploration is repeated $t_{1}$ times, and the final selected maximum and minimum observations (blue and yellow circles, respectively) are used as starting observations for the next iteration, in addition to adding $F_{i}\pm\epsilon_{i}$ (purple plane).}
	\label{fig:esquema_muce}
\end{figure}

In this Section, we propose the Multivariate Conditional Expectation (MUCE) method to tackle the main limitation of ICE (described in Section \ref{sec:ice}).
Since the ICE method is univariate, it is unaware of interactions between features that could influence the model's predictions.
MUCE is based on ICE curves, but it considers changes in multiple features simultaneously, thereby taking into account the changes that interactions between features could produce on the output. This method calculates the maximum and minimum changes in model output for the observation at study based on hill-climbing exploration \cite{russell2020aima} and therefore, in a greedy fashion. It explores the neighborhood of an observation by attempting to maximize and minimize the model's prediction after modifying any of the feature values. The difference between the maximum and minimum output predictions is an indicator of the model's output variation, which can be interpreted as the possible uncertainty of the model.
Figure~\ref{fig:esquema_muce} gives an overview of the method.

More in detail, the MUCE algorithm explores the neighborhood of a given observation while fixing a feature of interest $F_{i} \in F$. In each iteration of the algorithm (each of the planes in Figure~\ref{fig:esquema_muce}), the neighborhood of the observation is explored through small increments and decrements ($\epsilon_{j}$) of the value of each feature $F_{j} \in F, i\neq j$.
Afterwards, we calculate predictions for these newly generated observations, and the maximum and minimum predicted values, along with their corresponding observations, are stored. 

This exploration is repeated $t_{1}$ times (typically, $t_{1}=5$) for the first iteration in order to extend the region of influence to a greater distance from the original observation. Further iterations will start from the previous selected maximum and minimum observations by modifying $F_{i}$ by $\pm\epsilon_{i}$ and proceed with the same kind of exploration using $t_{i}$ repetitions (typically, $t_{i}=1, i>1$).

Note that the exploration starts at the original observation and proceeds in two directions ($\pm\epsilon_{i}$) and, for each direction, the exploration continues with the last chosen observation with the highest and lowest values of $\hat{y}$ (the predicted output), respectively. Thus, despite starting from an individual observation, the exploration continues for up to four generated observations (depending on draws). Algorithm \ref{code:muce} shows the pseudocode of the method.

\begin{algorithm}[!hbtp]
\caption{Pseudocode for MUCE}
\label{code:muce}
\begin{algorithmic}
	\Function{GenerateCandidates}{$\mathbf{x}, C, \boldsymbol{\epsilon}$}
		\State $X_{cand} \gets \emptyset$
		\For{$col \in C$}
			\State $\mathbf{x}_{new} \gets \mathbf{x}$
			\State $\mathbf{x}_{new}[col] \gets \mathbf{x}[col] \pm \boldsymbol{\epsilon}_{col}$
			\State $X_{cand} \gets X_{cand} \cup \{\mathbf{x}_{new}\}$
		\EndFor
		\State \textbf{return} $X_{cand}$
	\EndFunction

    \Function{PerformMuceSearch}{$\mathbf{x}, F_{i}, method, N, \mathbf{nsteps}, \boldsymbol{\epsilon}$}
    \Comment{$\mathbf{x}$ observation to test, $F_{i}$ feature of interest, $method$ max or min, $N$ number of iterations, $\mathbf{nsteps}$ number of repetitions for exploration during each iteration, $\boldsymbol{\epsilon}$ increments or decrements for each feature}

    \State $C \gets columns(\mathbf{x}) - F_{i}$
    \State $\mathbf{x}_{best} \gets \mathbf{x}$ 
    \State $R_{x} \gets \emptyset$ \Comment{Result observations}
    \State $R_{\hat{y}} \gets \emptyset$ \Comment{Result predictions for observations in $R_{x}$}
    \State $n \gets 0$
    \For{$n < N/2$}
	    \Comment{$N/2$ since at each iteration $F_{i}$ is incremented and decremented by $\epsilon_{i}$}
        \State $k \gets 0$
        \For{$k < \mathbf{nsteps}[n]$}
          	\State $\mathbf{x}_{k} \gets \mathbf{x}_{best}$
            \State $X_{cand} \gets GenerateCandidates(\mathbf{x}_{k}, C, \boldsymbol{\epsilon})$
            \State $\mathbf{\hat{y}} \gets model(X_{cand})$
            \State $\hat{y}_{k} \gets method(\mathbf{\hat{y}})$
            \If{$IsBetter(\hat{y}_{k}, \hat{y}_{best})$}
            	\Comment{If $\hat{y}_{k}$ is better than $ \hat{y}_{best}$ depending on \textit{method}}
	            \State $\hat{y}_{best} \gets \hat{y}_{k}$
	            \State $\mathbf{x}_{best} \gets X_{cand}[l] \mid \mathbf{\hat{y}}[l] = \hat{y}_{best}$
	            \Comment{$l$ is the index such that $\mathbf{\hat{y}}[l] = \hat{y}_{best}$}
	            \State $k \gets k+1$
            \Else
            	\State \textbf{break}
            \EndIf
        \EndFor
        \State $R_{x} \gets R_{x} \cup \mathbf{x}_{best}$
        \State $R_{\hat{y}} \gets R_{\hat{y}} \cup \hat{y}_{best}$
   	    \State $\mathbf{x}_{best}[F_{i}] \gets \mathbf{x}_{best}[F_{i}] \pm \epsilon_{i}$
        \State $n \gets n+1$
    \EndFor
    \State \textbf{return} $R_{x}, R_{\hat{y}}$
\EndFunction\\
\end{algorithmic}

\end{algorithm}

The algorithm outputs two sets of observations, along with their respective predictions. One set represents the maximum $\hat{y}$ observed during the exploration ($R_{\hat{y}}^{M}$, also known as ``MUCE max''), and the other represents the minimum ($R_{\hat{y}}^{m}$, also known as ``MUCE min''). These sets can be plotted to visualize the evolution of the prediction with respect to $F_{i}$.
The area between these two curves provides an idea of the prediction uncertainty, as it represents the changes resulting from minor variations in the input features.

Additionally, the maximum and minimum calculated $\hat{y}$ can be extracted from these sets, along with the corresponding values needed to achieve them. These feature variations can be used to gain additional insights into the model's behavior.

\subsection{Stability range}
\label{sec:stability_range}

To focus the analysis on a local region and provide local explainability, we define the stability range as the region around the observation values for which we want to evaluate model changes. More formally, for a feature $F_i$, we define the stability range $S_{i}$ as $S_{i}=[\mathbf{x}[F_{i}]-\delta_{i}, \mathbf{x}[F_{i}]+\delta_{i}]$, where $\mathbf{x}$ is an observation. 
The stability range must be linked to the intrinsic variability of each feature; therefore, each feature may have its own value for $\delta$. In a real setting, $\delta_{i}$ could be the known measurement error of a sensor. A too narrow stability range has the potential to obscure a substantial variation in the model's prediction, leading to an optimistic confidence. Conversely, an excessively broad stability range can increase the likelihood of a prediction fluctuation, thereby diminishing confidence.

ICE curves can be calculated only within the range $S$, limiting predictions to the local vicinity of the observation. In MUCE, $\epsilon_{i}$ depends on $\delta_{i}$ and the number of iterations ($N$): $\delta_{i}=\epsilon_{i}N$. Additionally, the observation generation for each feature $F_{j}$ is also limited to $S_{j}$, so all feature variations are confined to their stability range. In Figure~\ref{fig:esquema_muce}, the size of each plane (iteration) will be determined by $\mathbf{x}[F_{i}]\pm\delta_{i}$. By limiting the exploration in both methods, we focus on the local vicinity of the original observation and save computational time.

\subsection{Confidence indices in model prediction: stability and uncertainty}
\label{sec:metrics}

By using the methods from the previous Sections, several plots providing visual summary information can be generated for each feature using the methods described above. However, individually analyzing them can become tedious, especially as the number of features increases. To overcome this drawback, we propose a set of indices that summarize information within the stability range (as defined in Section~\ref{sec:stability_range}), enabling quantitative comparisons across features. These indices are primarily designed for classification tasks, although they are not limited to this purpose. In this context, we can assume that the output a posteriori probability is constrained to a real value between 0 and 1.

Stability quantifies the extent to which the model's output changes when only the feature of interest is modified (equation~\ref{eq:stability}). Higher values indicate more stable model predictions. Note that, in this context, we consider only $\hat{y}$ as obtained from the ICE curve inside the stability range.

\begin{equation}
\label{eq:stability}
stability=1-(max(\hat{y}_{ice})-min(\hat{y}_{ice}))
\end{equation}

Global uncertainty quantifies the extent to which the model's output varies when other features are also modified within the range of interest (Equation~\ref{eq:uncertainty}). It is calculated as the average difference between the maximum and minimum MUCE curves across all iterations ($N$). Higher values reflect greater divergence between these extremes, and therefore, increased model uncertainty. Such high uncertainty may suggest that the observation lies in a region the model has learned as a decision boundary between classes.

\begin{equation}
\label{eq:uncertainty}
    uncertainty=\frac{\sum_{i=-N/2}^{N/2}(R_{\hat{y},i}^{M} - R_{\hat{y},i}^{m})}{N}
\end{equation}

Uncertainty can be analyzed by distinguishing between decreasing (\textit{minus}) and increasing (\textit{plus}) the value of the feature of interest. For instance, a high value of $uncertainty^{+}$ combined with a low $uncertainty^{-}$ value indicates that the model's uncertainty is higher when the value of the feature of interest increases compared to when it decreases. Equations~\ref{eq:uncertainty_plus} and \ref{eq:uncertainty_minus} formally define these indices.

\begin{equation}
	\label{eq:uncertainty_plus}
	uncertainty^{+}=\frac{\sum_{i=0}^{N/2}(R_{\hat{y},i}^{M} - R_{\hat{y},i}^{m})}{N/2}
\end{equation}

\begin{equation}
	\label{eq:uncertainty_minus}
	uncertainty^{-}=\frac{\sum_{i=0}^{-N/2}(R_{\hat{y},i}^{M} - R_{\hat{y},i}^{m})}{N/2}
\end{equation}

\subsection{Using ICE and MUCE with non-numerical data}
\label{sec:other_data_types}

The previous Sections describe the methods under the assumption that the data consist of non-integer real numbers. However, real-world structured datasets typically contain heterogeneous data that have certain restrictions, such as categorical features. This Section describes how to treat different data types.

On the one hand, when dealing with integer features, the extreme values of the stability range must be rounded to the nearest integer. Parameter $\epsilon$ is also rounded, with the constraint $\epsilon \ge 1$.
The same adjustments apply to ordinal and binary features, as they are numerically represented as integers.
Note that it is impossible to define a stability range in binary features since there are only two possible values. Consequently, both values are always evaluated.

On the other hand, unlike the other types of features considered, non-ordered categorical features require a distinct approach. To implement the methods outlined in this work, it is necessary to determine which categories are most likely to be nearer to the observation under study. This prioritization of categories is particularly useful when dealing with a categorical feature with high cardinality, in order to give priority to the most probable categories around the observation value. This sorting method could include expert knowledge, being ad hoc for this specific categorical feature. For instance, if the categories represent cities, the sorting method could return the $k$ cities closest to the observation. In this work, we propose an approach based on the average distance to the $k$ nearest neighbors for each category from the observation of interest, and sort the categories according to this distance. For this distance calculation, all features in the dataset are used, except for the categorical feature to evaluate. The value of this categorical feature is used as the target. Note that in the presence of more than one categorical feature, this process is repeated for each of these features, so one categorical feature can participate in the computation of the distance of a second one.
The number of final selected values for each categorical feature depends on the stability range, taking into account the total number of values for the feature, rather than the range used for numerical features. In all cases, at least three categories are always considered, including the one associated with the original observation.

\section{Results}
\label{sec:results}

This section presents the results obtained by the explainable AI methods proposed in this paper. For this purpose, the datasets described in Section \ref{sec:dataset} were used to train XGBoost models \cite{xgboost} and then tested to assess their aptness for interpretability analysis. Their performance is summarized in Section \ref{sec:model-performance}. In the remainder of this section, the methods described in this work are applied to generate explanations for each model and observation of interest. 
For all the experiments, we define the stability range introduced in Section~\ref{sec:stability_range} as 5\% of the range of each numerical feature, $\delta_{i}=0.05(max(F_{i})-min(F_{i}))$. All values are considered for categorical and ordinal features, due to the low cardinality of these features in our dataset.

\subsection{Model performance}
\label{sec:model-performance}

Table \ref{tab:model_performance} shows the performance of the trained XGBoost classifiers for each dataset described in Section \ref{sec:dataset}. All models were evaluated using a random sampling of 70\% of the data for training and 30\% for testing, with the area under the ROC curve (AUC) as the primary evaluation metric.

\begin{table}[hbtp]
	\centering
	\caption{Performance of XGBoost models on the different datasets.}
	\label{tab:model_performance}
	\begin{tabular}{llc}
		\hline
		\textbf{\textit{Dataset}} & \textbf{\textit{Description}} & \textbf{\textit{ROC AUC} (\%)} \\
		\hline
		Synthetic 2D     & Linearly separable, 2 features      & 100.00 \\
		Synthetic 3D     & Non-linear boundary, 3 features     & 99.87 \\
		Real-world (original)   & Original features (float, int.) & 96.77 \\
		Real-world (transformed) & Mixed feature types (float, int., bin., ord., cat.) & 87.55 \\
		\hline
	\end{tabular}
\end{table}

As shown, the models trained on the synthetic datasets achieved nearly perfect classification, indicating a clear separation between classes. The model trained on the original real-world dataset also achieved high performance (96.77\% AUC), confirming its suitability for predictive modeling.
As expected, the performance decreased slightly after applying feature transformations to the real-world dataset. This reduction is likely not due to increased dataset complexity, but rather to the loss of granularity in the original continuous features. By converting features into binary, ordinal, and categorical formats, the model has access to less detailed information, which may limit its ability to capture subtle patterns. Nevertheless, the model still achieved a robust AUC of 87.55\%, confirming that the transformed dataset remains suitable for interpretability analysis.

\subsection{Model behavior in response to individual variations}
\label{sec:results-ice}

Using the algorithms described in Section \ref{sec:ice}, we can extract valuable insights, such as identifying the feature that induces the most significant individual change in the model output within the context of a given observation, while keeping all other features constant. This insight is achieved by analyzing the stability index and selecting the feature with the lowest stability score.

For illustrative purposes, we consider a model trained on a synthetic 2D dataset. Section \ref{sec:dataset_2d} describes the dataset and the representative observations selected to support the interpretation of the model behavior and the stability index. The feature values for observations TP0, TP1, and TP2, along with the corresponding indices computed for each feature, are presented in Table \ref{tab:stability_metrics}. 

\begin{table}[hbtp]
	\centering
	\caption{Feature values and stability indices for the three selected observations in the 2D synthetic dataset.}
	\begin{tabular}{lccc}
		\toprule
		\textbf{\textit{Point}} & \textbf{\textit{Feature}} & \textbf{\textit{Value}} & \textbf{$\boldsymbol{stability}$} \\
		\midrule
		\multirow{2}{*}{TP0} & \textit{F1} & 0.81 & 1.00 \\
		& \textit{F2} & -0.12 & 0.06 \\
		\midrule
		\multirow{2}{*}{TP1} & \textit{F1} & 1.16 & 0.34 \\
		& \textit{F2} & 0.33 & 0.84 \\
		\midrule
		\multirow{2}{*}{TP2} & \textit{F1} & -0.87 & 1.00 \\
		& \textit{F2} & 1.30 & 1.00 \\
		\bottomrule
	\end{tabular}
	\label{tab:stability_metrics}
\end{table}

\begin{figure}[hbtp]
	\centering
	\includegraphics[width=1\textwidth]{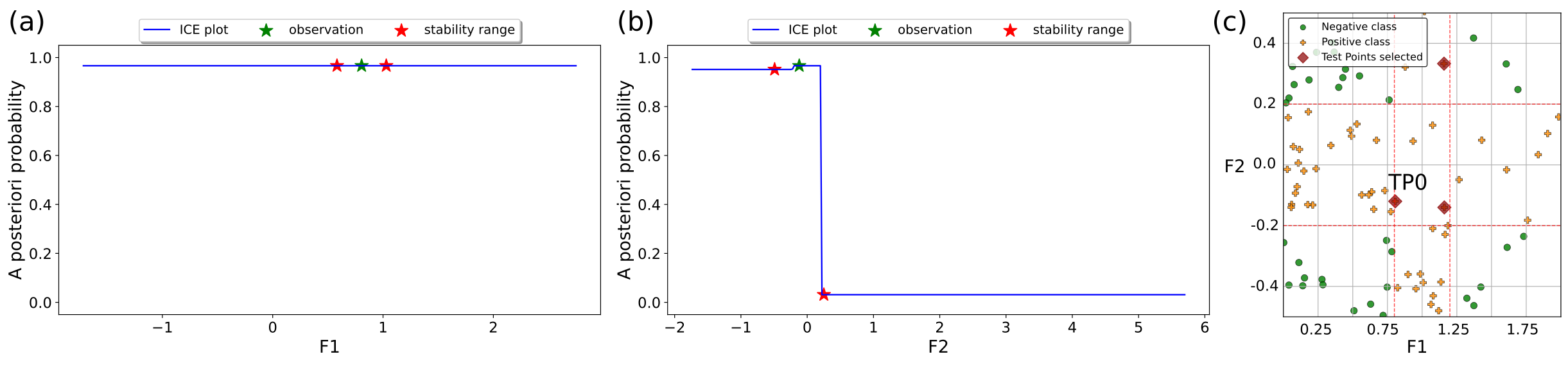}
	\caption{ICE curves for features \textit{F1} (a) and \textit{F2} (b) corresponding to the selected observation TP0 (green star). The red stars indicate the stability range (5 \%). Subfigure (c) shows a zoomed-in view of the synthetic 2D dataset centered around TP0.}
	\label{fig:ICE_TP0}
\end{figure}

The analysis of observation TP0 reveals that \textit{F2} is highly unstable (its stability value is close to zero), and, consequently, this feature may cause significant changes in the model prediction. This behavior is illustrated by the ICE curve corresponding to \textit{F2} in Figure \ref{fig:ICE_TP0} (b). Note that the \emph{a posteriori} probability within the region defined by the stability range varies greatly (from 0.1 to 0.9), indicating poor stability.

The observation TP0 is interesting for two reasons. On the one hand, it lies near a decision boundary, being correctly classified as a positive observation. When the value of \textit{F2} decreases, the \emph{a posteriori} probability drops only slightly because, as illustrated in Figure \ref{fig:dataset_synthetic2D}, despite the observation moves towards the boundary with the negative class, the high density of positive observations in that region allows the model to maintain a correct prediction.
On the other hand, increasing the value of \textit{F2} moves the observation near the negative class boundary, but in a sparsely populated area. As a result, the model produces an incorrect class prediction, influenced by the sparsity and structure of the local training data. This behavior is effectively captured by the stability index, which, with a value close to zero, already signals that the model output is unstable. 

For the feature \textit{F1}, the analysis reveals perfect stability (a stability index of 1), as clearly illustrated in Figure \ref{fig:ICE_TP0} (a), where varying the value of this feature does not affect the model's prediction. Because TP0 is surrounded by other observations consistently belonging to the positive class, the \emph{a posteriori} probability remains unchanged, indicating strong robustness in this area of the feature space.

\begin{figure}[H]
	\centering
	\includegraphics[width=1\textwidth]{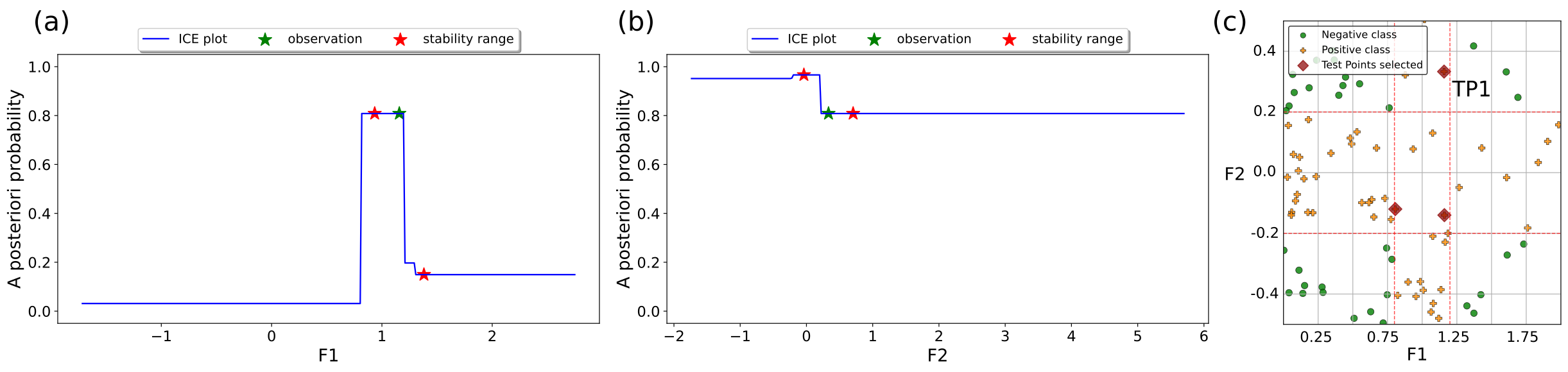}
	\caption{ICE curves for features \textit{F1} (a) and \textit{F2} (b) corresponding to the selected observation TP1 (green star). The red stars indicate the stability range (5 \%). Subfigure (c) shows a zoomed-in view of the synthetic 2D dataset centered around TP1.}
	\label{fig:ICE_TP1}
\end{figure}

The analysis of observation TP1, which is also located in a decision boundary region (see Figure~\ref{fig:dataset_synthetic2D}), exhibits relatively low stability for the feature \textit{F1}. If the value of \textit{F1} is slightly increased, the observation crosses into the negative class region, resulting in a class change based on its \emph{a posteriori} probability. In contrast, if the value is slightly decreased, it remains inside the region identified as positive, and the \emph{a posteriori} probability remains unchanged (see Figure \ref{fig:ICE_TP1} (a)). This behavior is effectively captured by the stability index because, although the model decision is correct, the proximity to the decision boundary introduces a certain degree of instability. 

When varying the value of \textit{F2} for TP1, the observation consistently remains within a region classified as positive. The model maintains a clear and stable prediction, as reflected in both the posterior probability (Figure \ref{fig:ICE_TP1} (b)) and the stability index. The slight decrease in posterior probability observed when increasing the value of \textit{F2} is due to the observation entering a less populated region of the feature space. Nonetheless, both the index and the ICE curve confirm that the model prediction remains stable in this area. 

Regarding observation TP2, it is located in a densely populated region by instances with the negative class and far from the decision boundary. These facts result in complete stability in both features, with a stability index equal to 1.00.

\begin{figure}[H]
	\centering
	\includegraphics[width=0.9\textwidth]{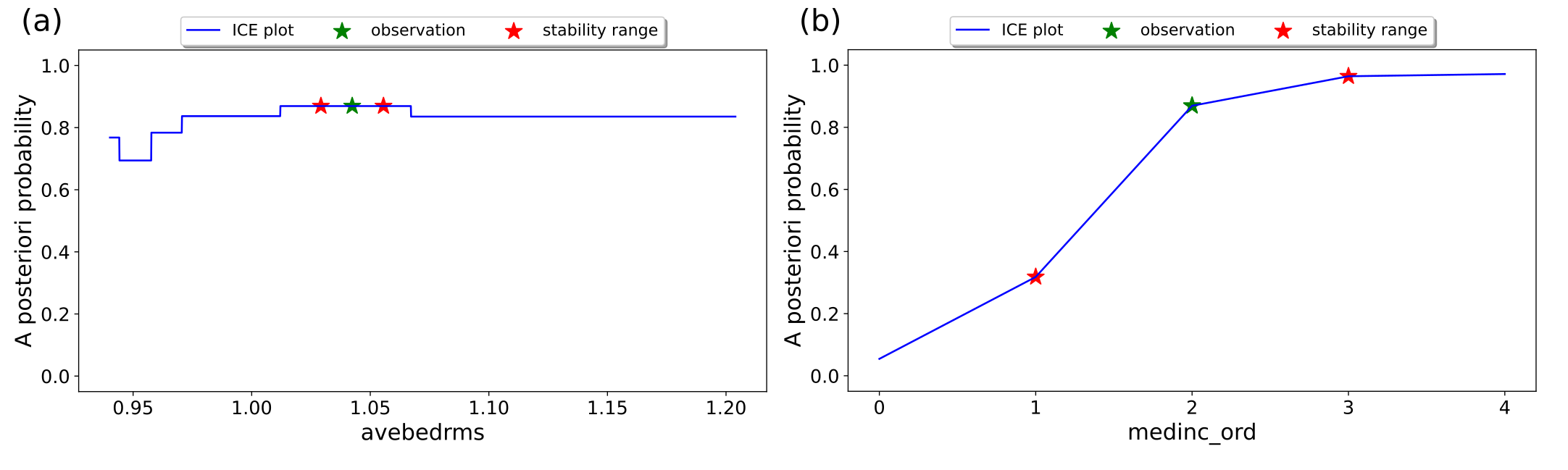}
	\caption{ICE curves for features \textit{avebedrms} (a) and \textit{medinc\_ord} (b) corresponding to an observation from the real-world dataset (green star). The red stars indicate the stability range (5 \%).}
	\label{fig:ICE_real}
\end{figure}

To further demonstrate the applicability of the proposed methodology, we now turn to a real-world dataset. While the previous analysis focused on synthetic data to highlight the behavior of the model under controlled conditions, this real-world dataset provides complementary examples drawn from a more realistic context. The focus remains on understanding how individual feature variations influence the model output, as captured by the stability index. In this context, two features from a selected observation were chosen to illustrate contrasting behaviors (Table \ref{tab:real_stability_metrics}). The feature \textit{avebedrms} exhibits perfect stability, with a stability index of 1.0 (Figure \ref{fig:ICE_real} (a)), indicating that the model output remains unaffected by perturbations in this feature. In contrast, the ordinal feature \textit{medinc\_ord} displays the lowest stability among all features analyzed, with a stability index of 0.35. For the selected observation, varying \textit{medinc\_ord} results in significant changes in the posterior probability, with a variation of approximately 65\% (Figure \ref{fig:ICE_real} (b)). This result clearly illustrates how local sensitivity to a particular feature can lead to potentially unstable or unreliable predictions.

\begin{table}[hbtp]
	\centering
	\caption{Feature values and stability indexes for a selected observation from the real-world dataset.}
	\label{tab:real_stability_metrics}
	\begin{tabular}{lcc}
		\hline
		\textbf{\textit{Feature}} & \textit{\textbf{Value}} & \textbf{$\boldsymbol{stability}$} \\
		\hline
		\textit{houseage}        & 51    & 0.98 \\
		\textit{averooms}        & 5.23  & 0.99 \\
		\textit{avebedrms}       & 1.04  & 1.00 \\
		\textit{aveoccup}        & 2.84  & 0.99 \\
		\textit{population\_bin} & 0     & 1.00 \\
		\textit{medinc\_ord}     & 2     & 0.35 \\
		\textit{cardinal\_point} & SW    & 0.55 \\
		\hline
	\end{tabular}
\end{table}

\subsection{Model behavior in response to combinations of variations}

\begin{figure}[hbtp]
	\begin{centering}
		\includegraphics[width=1\textwidth]{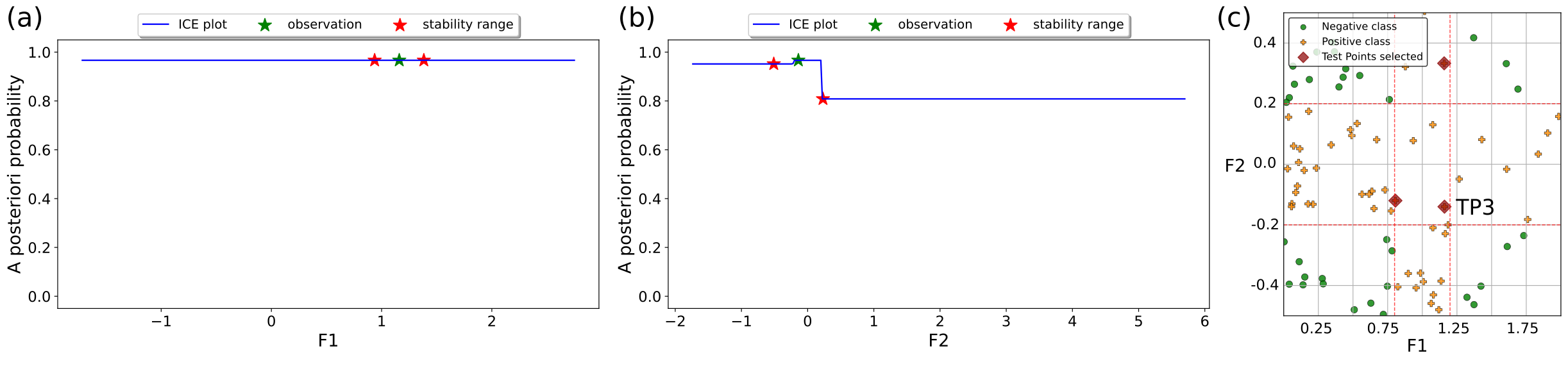}
		\caption{ICE results for TP3 of the synthetic 2D dataset, showing (a) feature \textit{F1} and (b) feature \textit{F2}. Subfigure (c) shows a zoomed-in view of the synthetic 2D dataset centered around TP3.}
		\label{fig:tp3_2d_ice}
	\end{centering}
\end{figure}

Sometimes, changes in model output are not attributable to variations in a single feature, but rather to a combination of variations in several features. In such circumstances, ICE is likely to be unable to detect these multivariate changes, whereas MUCE has been developed to address this scenario. We chose the observation TP3 of the 2D dataset presented in Section~\ref{sec:dataset_2d} to illustrate it. The decision for the positive class of this feature is not modified when \textit{F1} or \textit{F2} are varied in isolation, as shown in Figure~\ref{fig:tp3_2d_ice}, where only the ICE curve of \textit{F2} detects a slight change, although not sufficient to alter the predicted class.

\begin{figure}[hbtp]
	\begin{centering}
		\includegraphics[width=1\textwidth]{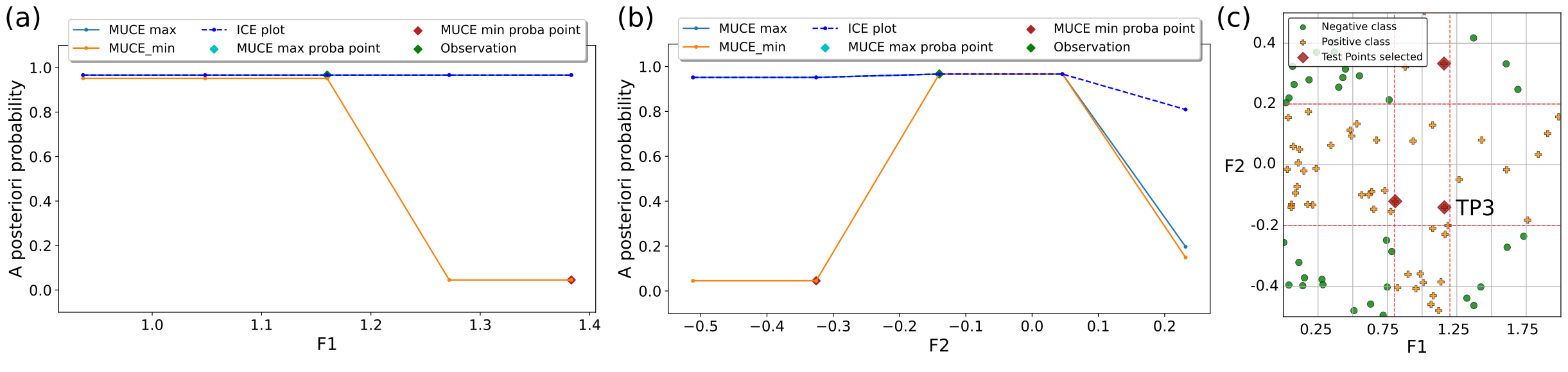}
		\caption{MUCE results for TP3 of the synthetic 2D dataset for (a) feature \textit{F1} and (b) feature \textit{F2}. The solid blue and yellow lines correspond to $R_{\hat{y}}^{M}$ and $R_{\hat{y}}^{m}$, respectively. The blue dashed line represents the ICE curve within the stability range. The estimated maximum and minimum probabilities are depicted in maroon and cyan dots; (c) shows a zoomed-in view of the synthetic 2D dataset centered around TP3.}
		\label{fig:tp3_2d_muce}
	\end{centering}
\end{figure}

However, variations in both features could bring the observation closer to the negative class. The results of MUCE are illustrated in Figure~\ref{fig:tp3_2d_muce}, where the results of considering \textit{F1} and \textit{F2} as features of interest are shown in the graphs (a) and (b), respectively. In the case of varying \textit{F1} in conjunction with \textit{F2}, the \emph{MUCE minimum curve} reflects a sharp change in the output probability when \textit{F1} is increased.
When varying \textit{F2}, a change is also detected, both when incrementing and decrementing the feature in combination with the others (Figure~\ref{fig:tp3_2d_muce}\,(b)). In this latter case, note that the maximum MUCE curve (cyan) is similar to the minimum curve, and the ICE curve (dashed blue) has a greater value. This behavior can be attributed to the greedy nature of the algorithm, which caused it to reach a local maximum. During the exploration phase, ties in output probabilities may occur, which in subsequent iterations can lead the process to converge toward a local minimum or maximum, limiting the effectiveness of the search.

The numerical value of stability and uncertainty indices is shown in Table~\ref{tab:tp3_2d_indicadores}. The stability is high for both \textit{F1} and \textit{F2} features, although slightly lower for \textit{F2}, as suggested by the ICE plots of Figure \ref{fig:tp3_2d_ice}. Uncertainty scores provide the user with an indication of how sensitive the predicted probability is to perturbations in the input features. Thus, complete range uncertainty is low; however, $uncertainty^{-}$ and $uncertainty^{+}$ are high for \textit{F2} and \textit{F1}, respectively, indicating that the predicted probability is more sensitive when \textit{F1} increases, as well as when \textit{F2} decreases.

\begin{table}[hbtp]
	\centering
	\caption{Indices summary for observation TP3 of the synthetic 2D dataset.}
	\label{tab:tp3_2d_indicadores}
	\begin{tabular}{lcccc}
		\toprule
		\textbf{\textit{Feature}} & \textbf{$\boldsymbol{stability}$} & \textbf{$\boldsymbol{uncertainty}$} & \textbf{$\boldsymbol{uncertainty^{-}}$} & \textbf{$\boldsymbol{uncertainty^{+}}$} \\
		\midrule
		\textit{F1} & 1.00 & 0.38 & 0.01 & 0.62 \\
		\textit{F2} & 0.84 & 0.38 & 0.60 & 0.02 \\
		\bottomrule
	\end{tabular}
\end{table}

The values of the features that produce the more significant changes in the model output can be presented as a complementary result. Thus, the observations with the highest and lowest predicted \emph{a posterior} probability can be easily extracted from the MUCE curves, and the feature variation magnitudes (FV) to be applied to the original observation may also be derived. These alternative observations could act as counterfactuals to the original observation.
In Figure~\ref{fig:tp3_2d_muce}, the maximum and minimum estimated probabilities are shown in marron and cyan, respectively. Note that, in this example, the maximum probability is already achieved by the original observation. Table~\ref{tab:tp3_2d_feat_variation} shows the FV values of \textit{F1} and \textit{F2} related to the minimum probability observed. Both cases require increasing \textit{F1} by 0.22 and decreasing \textit{F2} by 0.19.

\begin{table}[hbtp]
	\centering
	\caption{Feature variation (FV) for the estimated minimum probability of the observation TP3 of the synthetic 2D dataset. Each cell represents the change in each feature (rows) in the case of MUCE, computed for each feature (columns).}
	\label{tab:tp3_2d_feat_variation}
	\begin{tabular}{lccc}
		\toprule
		\textbf{\textit{Feature}} & \textbf{\textit{Value}} & \textbf{\textit{FV in F1}} & \textbf{\textit{FV in F2}} \\
		\midrule
		\textit{F1} & 1.16 & 0.22 & 0.22 \\
		\textit{F2} & -0.14 & -0.19 & -0.19 \\
		\bottomrule
	\end{tabular}
	
\end{table}

\begin{figure}[hbtp]
	\begin{centering}
		\includegraphics[width=1\textwidth]{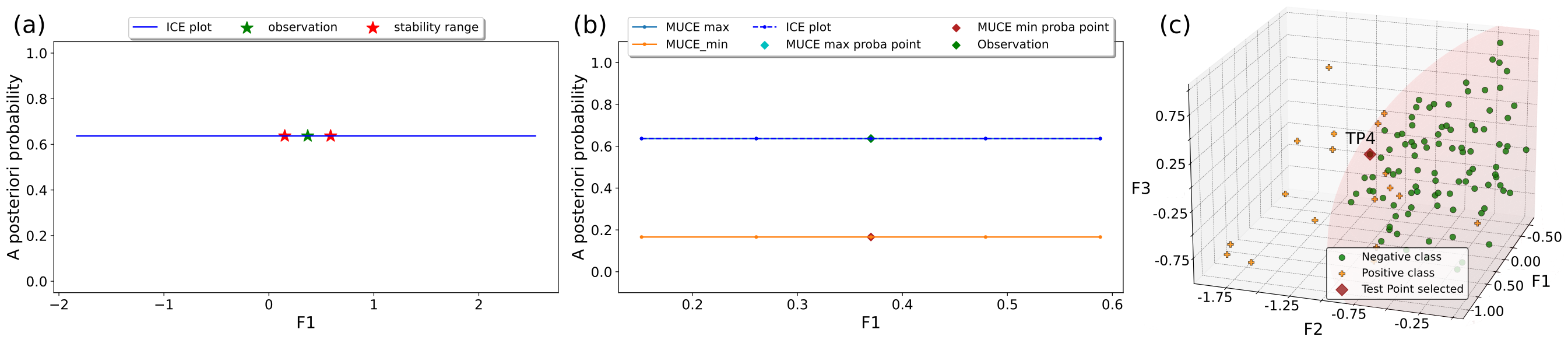}
		\caption{Results for feature \textit{F1} of the synthetic 3D dataset: (a) ICE and (b) MUCE. Subfigure (c) shows a zoomed-in view of the synthetic 3D dataset centered around TP4.}
		\label{fig:tp4_3d_ice_muce_f1}
	\end{centering}
\end{figure}

\begin{figure}[H]
	\begin{centering}
		\includegraphics[width=1\textwidth]{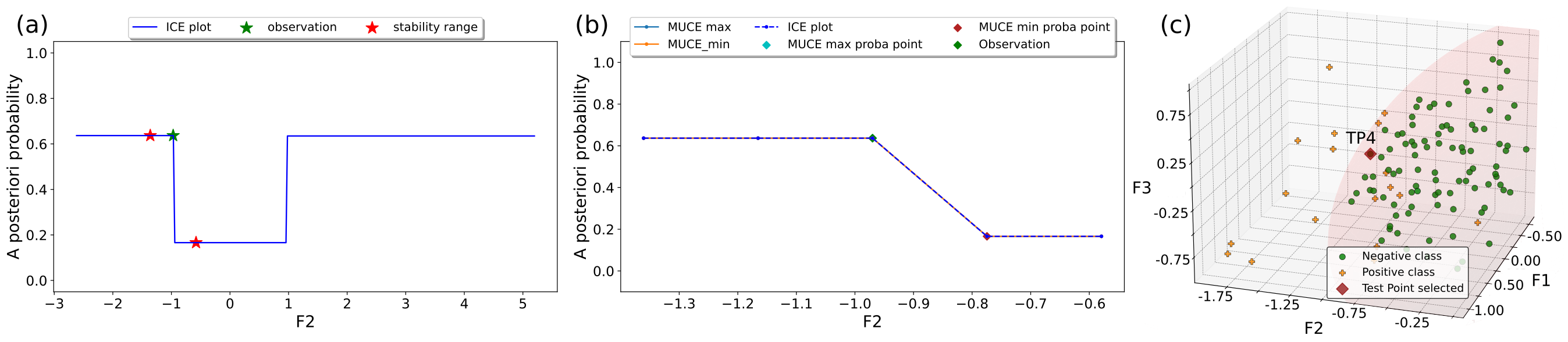}
		\caption{Results for feature \textit{F2} of the synthetic 3D dataset: (a) ICE and (b) MUCE. Subfigure (c) shows a zoomed-in view of the synthetic 3D dataset centered around TP4.}
		\label{fig:tp4_3d_ice_muce_f2}
	\end{centering}
\end{figure}

\begin{figure}[H]
	\begin{centering}
		\includegraphics[width=1\textwidth]{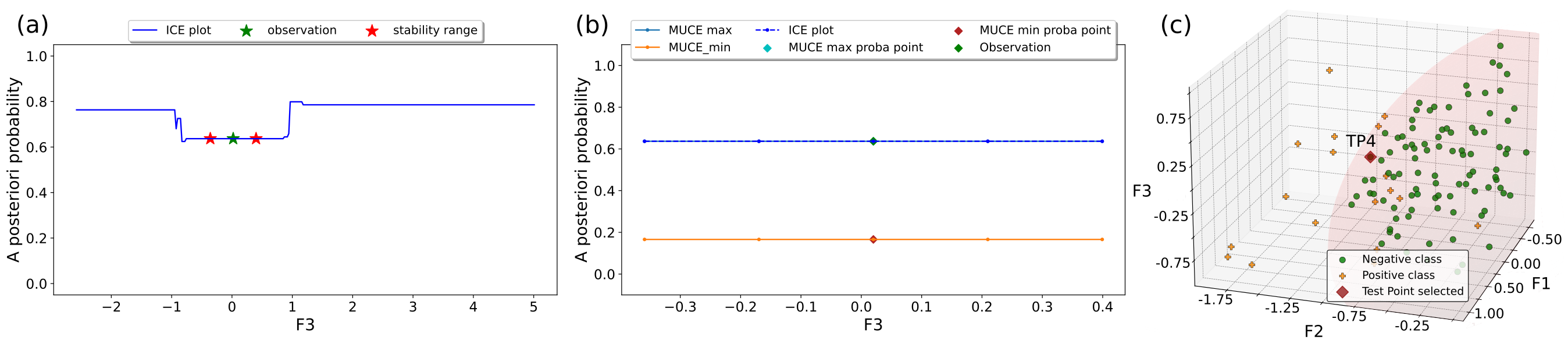}
		\caption{Results for feature \textit{F3} of the synthetic 3D dataset: (a) ICE and (b) MUCE. Subfigure (c) shows a zoomed-in view of the synthetic 3D dataset centered around TP4.}
		\label{fig:tp4_3d_ice_muce_f3}
	\end{centering}
\end{figure}

To test the algorithms presented in this work on a nonlinear dataset and a higher number of features, we analyzed observation TP4 from the 3D dataset. The results on stability and uncertainty are shown in Table~\ref{tab:tp4_3d_indicadores}. The stability indexes obtained for the features \textit{F1} and \textit{F3} were very high while the uncertainty was moderate, as illustrated in Figures~\ref{fig:tp4_3d_ice_muce_f1} and~\ref{fig:tp4_3d_ice_muce_f3}. In contrast, \textit{F2} exhibited low stability and minimal uncertainty, as illustrated in Figure~\ref{fig:tp4_3d_ice_muce_f2}, which suggests that a variation in \textit{F2} is necessary to induce a change in the model’s prediction.

\begin{table}[hbtp]
	\centering
	\caption{Indices summary for the observation TP4 of the synthetic 3D dataset.}
	\label{tab:tp4_3d_indicadores}
	\begin{tabular}{lccccc}
		\toprule
		\textbf{\textit{Feature}} & \textbf{\textit{Value}} & \textbf{$\boldsymbol{stability}$} & \textbf{$\boldsymbol{uncertainty}$} & \textbf{$\boldsymbol{uncertainty^{-}}$} & \textbf{$\boldsymbol{uncertainty^{+}}$} \\
		\midrule
		\textit{F1} & 0.37 & 1.00 & 0.47 & 0.47 & 0.47 \\
		\textit{F2} & -0.97 & 0.53 & 0.00 & 0.00 & 0.00 \\
		\textit{F3} & 0.02 & 1.00 & 0.47 & 0.47 & 0.47 \\
		\bottomrule
	\end{tabular}
\end{table}

Regarding feature variations, for all features, the maximum \emph{a posteriori} probability is achieved with the initial values of the observation. The results of Table~\ref{tab:tp4_3d_feat_variation} show FV related to the minimum \emph{a posteriori} probability. The results indicate that it is necessary to increase \textit{F2} by 0.20 to achieve the minimum estimated \emph{a posteriori} for MUCE centered in each feature. Assuming an \emph{a posteriori} threshold of 0.5, these changes will result in a change to the predicted class, since the estimated \emph{a posteriori} probability is around 0.2.

\begin{table}[H]
	\centering
	\caption{Feature variation (FV) for the minimum found a posteriori for TP4 of the synthetic 3D dataset. Each cell represents the change in each feature (rows) in the case of MUCE centered in the main feature (columns).}
	\label{tab:tp4_3d_feat_variation}
	\begin{tabular}{lccc}
		\toprule
		\textbf{\textit{Feature}} & \textbf{\textit{FV in F1}} & \textbf{\textit{FV in F2}} & \textbf{\textit{FV in F3}} \\
		\midrule
		\textit{F1} & -0.11 &    0 &    0 \\
		\textit{F2} &  0.20 & 0.20 & 0.20 \\
		\textit{F3} &  0.19 & 0.38 &    0 \\
		\bottomrule
	\end{tabular}
\end{table}

\subsection{Response to Variations in Discrete and Categorical Features}

To demonstrate the adaptability of the proposed approach to a broader range of data types, this section focuses on binary and categorical features. Although the methodological adaptations required to handle these kinds of features have already been described in Section \ref{sec:other_data_types}, it is important to analyze how variations on them affect the graphical representations used to interpret the model behavior. Table \ref{tab:real_stability_metrics} shows the indices for the selected observation. Note that the proposed indices depend on the predicted \emph{a posteriori}, so no adjustments are needed for other data types.

\begin{table}[H]
	\centering
	\caption{Stability and uncertainty values for the selected observation from the real-world dataset used in Section \ref{sec:results-ice}.}
	\label{tab:real-point_indicadores}
	\begin{tabular}{lcccc}
		\toprule
		\textbf{\textit{Feature}} & \textbf{$\boldsymbol{stability}$} & \textbf{$\boldsymbol{uncertainty}$} & \textbf{$\boldsymbol{uncertainty^{-}}$} & \textbf{$\boldsymbol{uncertainty^{+}}$}\\
		\midrule
		\textit{houseage}         & 0.98 & 0.87 & 0.87 & 0.87 \\
		\textit{averooms}         & 0.99 & 0.87 & 0.87 & 0.87 \\
		\textit{avebedrms}        & 1.00 & 0.87 & 0.87 & 0.87 \\
		\textit{aveoccup}         & 0.99 & 0.87 & 0.87 & 0.86 \\
		\textit{population\_bin}  & 1.00 & 0.87 & 0.87 & 0.87 \\
		\textit{medinc\_ord}      & 0.35 & 0.23 & 0.33 & 0.23 \\
		\textit{cardinal\_point}  & 0.55 & 0.80 & 0.80 & 0.80 \\
		\bottomrule
	\end{tabular}
\end{table}

\begin{figure}[H]
	\centering
	\includegraphics[width=0.9\textwidth]{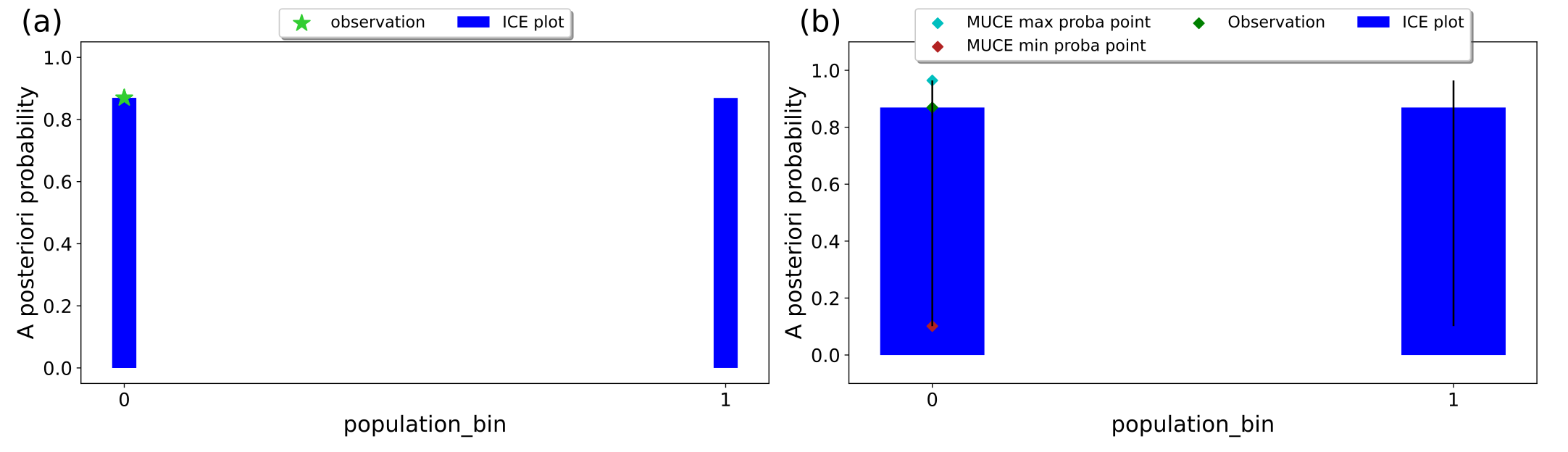}
	\caption{Visualization of ICE (a) and MUCE (b) curves for a binary feature.}
	\label{fig:real_binary}
\end{figure}

Integer and ordinal-valued features retain the same ICE and MUCE plot structure and interpretation as continuous-valued features since all of them follow a natural order.
In contrast, binary features are limited to two possible values and are represented using bar plots instead of continuous curves. To illustrate the graphs layouts and interpretations, we will use the feature \textit{population\_bin} of the selected observation from the real-world dataset as an example. The ICE plot of this feature is shown in Figure \ref{fig:real_binary} (a), where each bar corresponds to the \emph{a posteriori} probability of each binary value for this feature. Because the stability range does not apply to the case of binary features, the red stars used to denote the stability range boundaries are omitted.

Similarly to ICE, the MUCE plot of binary features can be drawn using a bar for each value with vertical lines superimposed on the bars to indicate the variation found by the algorithm in terms of estimated minimum and maximum \emph{a posteriori} probabilities. The indices of the feature \textit{population\_bin} are listed in Table \ref{tab:real-point_indicadores}, and its MUCE plot is shown in Figure \ref{fig:real_binary} (b). The plot depicts identical \emph{a posteriori} probabilities when no other features are changed. Despite this feature exhibiting perfect stability, uncertainty values are relatively high, as indicated by the length of the variation lines depicted on bars.

\begin{figure}[H]
	\centering
	\includegraphics[width=0.9\textwidth]{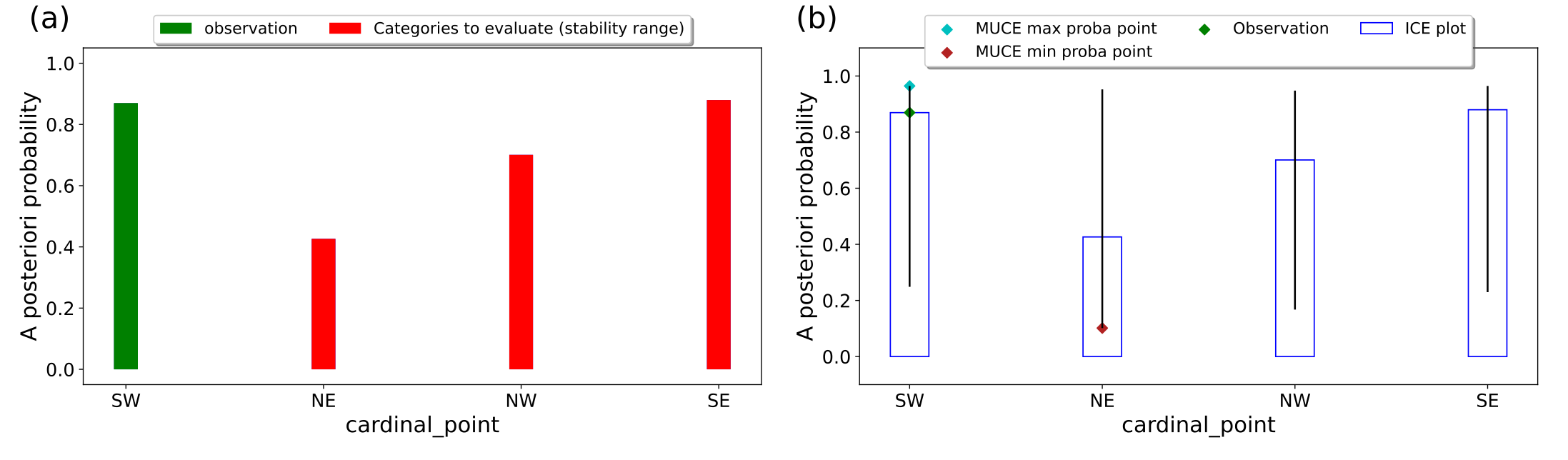}
	\caption{Visualization of ICE (a) and MUCE (b) curves for a categorical feature.}
	\label{fig:real_categorical}
\end{figure}

For categorical features, specific modifications are also required at the visualization level. As with binary features, bar plots are used to represent the ICE and MUCE outputs. In the ICE plot, the category corresponding to the observation under analysis is shown in green. In contrast, the closest categories, ordered according to the criterion defined in Section \ref{sec:other_data_types} (density and closeness to the observation of interest), are highlighted in red. 
In the MUCE plot, the height of each bar reflects the ICE value. At the same time, vertical lines indicate the range of posterior probabilities discovered by MUCE when varying the remaining features, as shown for binary features. We illustrate this behavior using the categorical feature \textit{cardinal\_point}. The stability index is 0.55, indicating moderate instability. In the ICE plot, the four categories present in the real dataset are considered within the stability range, with posterior probabilities ranging from just above 0.4 to nearly 0.9. This wide range reflects a substantial sensitivity to changes in the categorical input, and a possible class change is observed (Figure \ref{fig:real_categorical} (a)). Additionally, the model exhibits a high degree of uncertainty in this region (Table \ref{tab:real-point_indicadores})), which is clearly visible in the MUCE plot through the length of the vertical variation lines (Figure \ref{fig:real_categorical} (b)). 

\section{Discussion}
\label{sec:discuss}

This work presents a new local explainability technique called MUCE. This technique, combined with a modified version of the well-known ICE method, allows for the analysis of a model's response to small changes in the input features. The modification to ICE allows the visualization of results for a single isolated observation, a feature not typically supported by standard implementations. While ICE shows changes at the univariate level, MUCE considers variations at the multidimensional level. In the performed experiments, the models are explained through graphical outputs, as well as indices of stability and uncertainty in the model's output in the neighbourhood of a test observation.

To evaluate the performance of the explainability techniques presented, XGBoost models are trained on different datasets: two synthetic and one real-world. The synthetic datasets (2D and 3D) were meant to provide controlled environments for precise evaluation and to assess observations in diverse contexts, including edge cases such as observations located near the decision boundaries between classes. After being modified to introduce feature heterogeneity, the real-world dataset was used to adapt the methodology to different feature types and provide practical examples in a more realistic setting. As observed in the results (Section \ref{sec:results-ice}), the methodology effectively captures the model behaviour in complex regions, such as decision boundaries and sparsely populated areas. 
For instance, the analysis of specific observations, like TP0, successfully linked prediction instability (low stability index) to the model when it is moved into a sparsely populated area near the class boundary. As shown in the examples, a change in a single feature may be sufficient for the model’s output to shift significantly, even leading to a class change, according to the distribution of the dataset observations (the presence or absence of specific class observations in a region of the space).

However, in other scenarios, changes must occur in several features simultaneously to produce changes in the model's output. In these cases, the ICE methodology presents a limitation. Although it can be used to measure the effect of multiple feature changes, an exhaustive scan of all feature combinations is required to determine whether any of them produce a significant change when varied in combination. The proposed MUCE method overcomes this limitation, capturing multidimensional interactions without the need for an exhaustive search. The example of the observation TP3 on the synthetic 2D dataset illustrates this effect, as the interaction of two features is needed to produce a significant change in some regions of the space. In practice, both ICE and MUCE serve complementary roles: ICE is crucial for assessing the impact and stability of individual features. At the same time, MUCE explicitly handles the detection of necessary feature combinations that cause significant prediction shifts and quantify the model's uncertainty.

Although the main output of these methods is a graphical representation, this work introduces two indices --stability and uncertainty-- that allow the information to be summarised numerically and enable a rapid analysis without the need to inspect a large number of plots. In this sense, we provide two quantitative and objective indices, facilitating quantitative comparisons across features. 
Global uncertainty, being an average, may mask effects that occur only when the feature of interest is increased or decreased. To mitigate this effect, we propose splitting uncertainty into two components, $uncertainty^+$ and $uncertainty^-$, to obtain a more granular outcome. As seen in the analyzed examples, this allows for the detection of small changes at the local level, as well as changes occurring in a single direction within a locality.

The presented approach does not require training with additional models, as inference for the explanations can be computed on the target model, thereby overcoming some limitations of several approaches. For example, LIME assumes linearity in the surroundings, and other surrogated models could compromise the accuracy of the explanations by using an oversimplified model for a given task.

The proposed adaptation of ICE enables computation on a new observation at inference time thanks to the use of a grid of values to generate the surrounding observations. Furthermore, it has additional advantages. First, it is less sensitive to unpopulated regions of space. In the original ICE algorithm, the values of each feature needed to generate the curves have been observed previously on the dataset. In our adaptation, using a grid allows us to generate new observations even if the values have not been observed, and adjusting the step of the grid improves the precision of the curve. Secondly, the complete training data is not needed to generate the curves, as only the minimum and maximum values (or categories in the case of categorical data) are needed. Furthermore, the association between values for each observation is lost, which avoids the risk of reidentification in use cases where data is confidential, such as healthcare.

However, the presented approach has its own limitations. The sweep of the grid does not prevent the generation of unrealistic observations, especially in the presence of correlated values. Also, since MUCE follows a greedy approach, the results could reach a local maximum or minimum. Proposals can rely on extending the exploration in each iteration, or by introducing methodological refinements. It is also worth noting that the k-nearest neighbours approach for category sorting could imply some reidentification risk, and other approaches could be more appropriate for use cases where this issue is relevant.

Using stability and uncertainty indices as described cannot be directly used in regression tasks, survival analysis, or risk probability assessment, because their ranges must be properly delimited. The reason is that they assume a prediction estimation in terms of \textit{a posteriori} probability (ranging between 0 and 1), and the range of such predictions should be known in advance, which often is not possible. As future work, we aim to extend the use of such indexes to more types of tasks. Further limitations include the need for extensions to address non-tabular data such as images, sound, or text. 

A prospective line of research involves performing a comparative analysis with established methods from the literature, such as LIME and SHAP. This approach would enable the development of a hybrid explainability framework that integrates multiple techniques in a complementary fashion. Specifically, while LIME and SHAP focus on identifying relevant parameters, the proposed methodology could provide crucial information on the stability of those features, as well as the uncertainty the model presents when making that decision. This integration would thereby enhance the interpretability and trustworthiness of artificial intelligence models by providing enriched explanatory insights.

\section{Conclusions}
\label{sec:conclusions}

This work introduces a novel local explainability method, named MUCE, which aims to explain models in a multivariate manner. Additionally, an adaptation of ICE is presented that can be used at inference time on new test data, allowing the visualization of results for a single isolated observation. Two quantitative indices, stability and uncertainty, complement both ICE and MUCE methods, summarizing the behaviour observed through ICE and MUCE graphs, which enables the efficient analysis of numerous features. All methods are evaluated on both synthetic and real datasets to demonstrate their performance.

Future work includes extending these indices to other types of predictive tasks, such as regression. Moreover, it would be of interest to conduct a comparative study with established methods from the literature, such as LIME and SHAP. Overall, MUCE and the proposed ICE implementation may provide a complementary framework that enhances the interpretability of predictive models, offering both graphical and quantitative insights that can foster greater trust and understanding in AI systems.

\section*{Acknowledgements}

This work was funded by Generalitat Valenciana through IVACE (Valencian Institute of Business Competitiveness, \url{https://www.ivace.es/index.php/es/}) through a grant to support ITI's activity in independent research and development, research results dissemination and knowledge and technology transfer to business under projects IMAMCA/2024/11 and IMAMCA/2025/11. It was also funded by the Cervera Network for Leadership in R+D+I in Applied Artificial Intelligence (CEL.IA), co-funded by the Centre for Industrial and Technological Development, E.P.E. (CDTI) and by the European Union through the Next Generation EU Fund, within the Cervera Aids program for Technological Centres, with expedient number CER-20211022.

\bibliography{main}

\end{document}